# Applying Ontological Modeling on Quranic "Nature" Domain

A.B.M. Shamsuzzaman Sadi[1], Towfique Anam[2], Mohamed Abdirazak[3], Abdillahi Hasan Adnan[4], Sazid Zaman Khan[5], Mohamed Mahmudur Rahman[6], Ghassan Samara[7]

[1,2,3,4,5]Department of Computer Science and Engineering, International Islamic University Chittagong, Bangladesh
[6]Department of Science and Technology, Universiti Sains Islam Malaysia, Nilai, Malaysia
[7]Department of Internet Technology, Zarqa University, Jordan
Email: abmsadi06@gmail.com[1], towfiqueanam@protonmail.com[2], maxamed319@gmail.com[3], abdilaahi01@gmail.com[4], szkhanctg@gmail.com[5], provaiiuc@raudah.usim.edu.my[6], gsamara@zu.edu.jo[7]

*Abstract*— The holy Quran is the holy book of the Muslims. It contains information about many domains. Often people search for particular concepts of holy Quran based on the relations among concepts. An ontological modeling of holy Quran can be useful in such a scenario. In this paper, we have modeled nature related concepts of holy Quran using OWL (Web Ontology Language) / RDF (Resource Description Framework). Our methodology involves identifying nature related concepts mentioned in holy Quran and identifying relations among those concepts. These concepts and relations are represented as classes/instances and properties of an OWL ontology. Later, in the result section it is shown that, using the Ontological model, SPARQL queries can retrieve verses and concepts of interest. Thus, this modeling helps semantic search and query on the holy Quran. In this work, we have used English translation of the holy Quran by Sahih International, Protege OWL Editor and for querying we have used SPARQL.

## I. INTRODUCTION

Search Engines are gradually augmenting their search using semantic search technologies. For example, Google's Hummingbird algorithm is a major step towards semantic search on search engines. While semantic search is applicable for many domains, our goal is to search concepts, relations among concepts and verses of holy Quran using a semantic perspective. That means if one wants to search the verse that has concepts ALLAH and Rain and the relation between them; he should be able to get the desired result. Our final goal is a search engine that will be able to search any verse from holy Quran based on semantic or conceptual queries. But currently we are trying to build an ontology based on Quranic concepts as a foundation for this type of search engine. Since this is a work in progress and implementation of the semantic search engine has not yet been completed, we keep our discussions limited to methodologies of analyzing and searching holy Quran from an ontological perspective. This means that, for now, the search performance is not compared with familiar search engines available online; however this paper demonstrates a preliminary proof-of-concept and the basic principle for a semantic Quranic search engine. As holy Quran is a domain full of various subject matters, we are currently focusing on concepts that are related to Nature domain. Nature domain is a concept in holy Quran which has not yet been investigated extensively. The vision of a semantic web is extremely ambitious and would require solving many long-standing research problems in knowledge representation and reasoning, databases, computational linguistics, computer vision and agent systems. Semantic [1] search technology not only uses the keyword-based search but also makes relationship between two key concepts and the knowledge Base (KB). So the machine can understand the sentence using its knowledge about relationship between concepts that are in its KB. RDF triples are used to represent knowledge. This can help our users to search more accurately and precisely. Although, it has not been shown in this particular work, we are working towards a semantic search engine for holy Quran which is able to deal with search queries in colloquial English. Search made by users contain more concepts than keywords. To make it understandable to the machine and to find what exactly the user wants, semantic search is the best way for now. The first step of doing that is building a KB that will contain all the RDF triples that are in holy Quran. Human curiosity of knowing more about the nature and linking between nature and holy Quran leads them to ask critical questions. Often people can't put these queries in a simple keyword-based query format suitable for query-based search engines because of the lack of the knowledge of all keywords of the Quranic sentence. Actually semantic search mechanisms help in such situations. Rest of the paper is organized in the following way: Section II analyzes the literature, Section III discusses the methodology, Section IV shows the results, Section V presents the conclusion.

## II. LITERATURE REVIEW

Some works have been done in the field of Quranic information retrieval. Noorhan Hassan Abbas [2] on her thesis paper has done research about a tool and website that will help search a concept from holy Quran. As we know, the verses in holy Quran about a single topic is scattered over various chapters, it is quite hard to find everything related to particular concepts in holy Quran. She thus created a tool and a website to solve this problem which ultimately became Quranic Corpus project. Her work stands out as a distinguished work on information retrieval from holy Quran using semantic web technology. M A Sherif et al. [3] have worked with semantic Quran using Natural Language Processing techniques. They presented the Semantic Quran Dataset that includes 42 different language translations of holy Quran and they extracted the data from both the Tanzil Project [4] and Quranic Arabic Corpus [5]. They designed ontology by extracting data sources that consist of different language translations of holy Quran. Al-Zoghby et al. [6] presented a survey on Arabic Semantic web apps. They concluded that, the current web is much more than what we anticipated. Evolution of semantic web added another dimension to this problem and there is shortage of semantic web apps in Arabic. HU Khan et al. [7] described ontology based semantic search on their paper. According to them, it is quite difficult to implement semantic search for holy Quran. In some places some topics are explicitly mentioned and some are meant implicitly. They mainly covered animal domain in holy Quran and their work proposed the use of semantic web for semantic search in holy Quran so that it becomes easier to search topics efficiently. S Saad et al. [8] [9] [10] described development of ontology development for Islamic Literature. In [9] and [10], they presented an algorithm for automatic extraction of keywords while in [8] they presented a methodology to extract information from the Islamic knowledge to build an ontology for given domain.

ABM Sharaf et al. [11] investigated Quranic text mining. They have shown that there are actually texts about various topics in holy Quran which are scattered in various chapters of holy Quran. They created QurSim which is a resource for Quranic scholars, students and researchers who are going to pursue research on holy Quran. According to them, it is possible to expand their created dataset.

ZA Adhoni et al. [12] presented digital Quran API (Application Programming Interface) that will help create applications based on holy Quran. Among these two APIs are notable, The Quran API and the Quran Search API. The second one can be used to search holy Quran by a word or verse by using filters and also can be used to search for commentaries, transliterations and translation of holy Quran. This paves the way for developing data-rich applications for holy Quran.

J Dror et al. [13] presented a computation system for morphological analysis and annotation of holy Quran for research and teaching purposes. The system is useful for investigating several morphological, syntactic, semantic and stylistic aspects of the Quranic texts. P Saeedi et al. [14] made taxonomy of question types and ontology for the Quranic knowledge. Several other works [15-17] strived to develop ontological model of holy Quran, some of which details and demonstrates semantic storage and retrieval of Quranic concepts and verses of interest, while others do not. Use of linguistic tools is also noticeable in some of these works.

### III. METHODOLOGY

Before discussing our methodology, a brief discussion of Web Ontology Language (OWL) could be useful. Ontological modeling aims to enlist the concepts of a particular domain in a hierarchical manner so that superclass-subclass relations can be specified accurately. It also aims to specify the relations among these concepts. This is a form of knowledge representation which ultimately leads towards logic based formalism and reasoning about the concepts in the ontology. Indeed, the present version of OWL called OWL 2 is fundamentally built on Description Logic.

In order to model the holy Quran, we followed certain steps. We needed to model the concepts related to holy Quran using OWL/RDF. We divided the work into the following steps:

(1) Going through the whole holy Quran for several times. We have used the English translation of holy Quran by Sahih International [18] along with the Arabic holy Quran for our work purpose.
(2) Finding all of the related verses that are in relation with nature domain in holy Quran.
(3) Filtering Nature related verses, removing the verses which are hard to explain and where something related to nature is used as a metaphor.
(4) Creating RDF triples based on the verses that are left after filtering and then creating the ontology based on the RDF triples.
(5) Using SPARQL to query the ontology to test whether it can find the verses and concepts of interest. If it works, then turning the test ontology to fully fledged one else doing changes in RDF triples and doing this step again. Figure 1 shows the class hierarchy according to our ontology design.

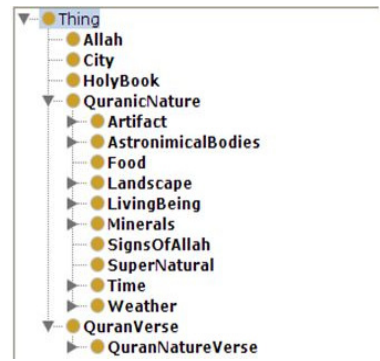

Fig. 1. Class hierarchy in the Quranic nature ontology.

We have read the holy Quran for several times to understand the meaning of those verses, collected our desired verses, filtered every time we read and tried to give cross check on those verses.

Then we made our triples according to those verses and by our analysis we divided each nature-related verse into concepts and the relationship between those concepts because triples are made with concepts and relationships. Then to implement our triple model, we have used OWL/RDF. For

OWL/RDF, we used Protégé OWL Editor [19][20][21]. We have added all the Concepts under OWL/Thing concept.

Under OWL/ Thing concept, we have created some sub classes (concepts) which are, for instance, Allah, City, HolyBook, QuranicNature, QuranVerse. Our main working classes are QuranicNature and QuranVerse. Under QuranicNature, sub classes are: Astronomical Bodies, Artifact, Food, Landscape, LivingBeing, Minerals, SignsofAllah, SuperNatural, Time, and Weather. These sub classes have more sub classes. QuranVerse contains the verses number as it's sub class. These verses have relationship with QuranicNature's sub classes.

We have put Allah, City, HolyBook out of the QuranicNature because of their specialty and these are not part of nature but they have relationship with QuranicNature and those have impact on the nature. To relate QuranicNature concept with QuranVerse we have used some inverse relationships (ObjectProperties) called hasPart and isPartOf. For example, In chapter 2, verse number 50 it is said: "And [recall] when we parted the sea for you and saved you and drowned the people of Pharaoh while you were looking on." So we have divided concepts and relationships and made triple as Figure 2.

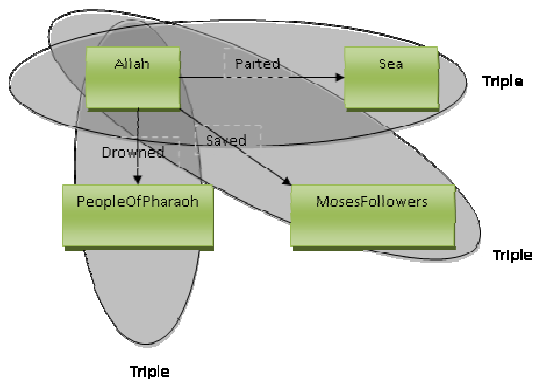

Fig. 2. Triple of verse 2:50.

We have taken strong elements as concepts (classes) and the connecting word between two elements as relationship (ObjectProperties). So this is just the basic knowledge triple but we have connected this knowledge with the verse/verse number using hasPart and isPartOf relationship as in Figure 3.

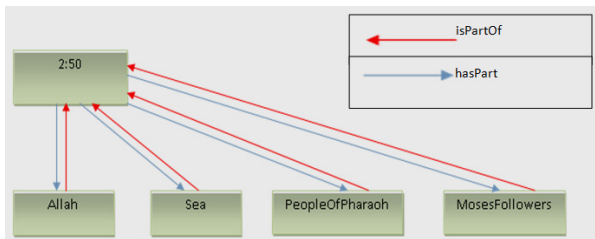

Fig. 3. Triple representation of verse 2:50.

## IV. RESULTS

To find out if the Ontology model is working properly or not, we had to run some tests using SPARQL query and see the output whether it is showing what we intend to show. We are using SPARQL-OWL to query from our ontology [22]. SPARQL is a query language to query RDF and OWL datasets. The main feature of SPARQL is query based on relations. Therefore, if the relation between two concepts and one concept among them are known, the unknown concept can be retrieved from the knowledge base. The result of SPARQL queries can be exported in several formats including JSON (Java Script Object Notation) and XML (Extensible Markup Language). Many ontology based domain-specific datasets contain public SPARQL endpoints on the web so that users can make queries to find particular relations and concepts of the corresponding domain.

By using SPARQL query we can make sure that, using our ontology we can retrieve relevant concepts and verses from the knowledge domain we have created using the holy Quran. Below are some sample queries we have created to test the ontology.

**Query 1:** Find the thing which Allah parted?
**SPARQL Query 1.** SELECT * WHERE{{qreg:Allah qreg:parted ?Answer.}}
**Answer:** Sea.

In query 1, we are selecting something unknown (?Answer). The query specifies the relationship of ?Answer with qreg: Allah as qreg: Parted. Here, qreg is our ontology prefix and Allah is the individual from the Allah concept. In other words, Allah is the subject in this query, parted is the predicate and ?Answer is our object.

**Query 2:** Find the verse which mentions Allah raised mountain?
**SPARQL Query 2.** SELECT ?Concept ?AyatNo ?Ayat WHERE {{qreg:Allah qreg:raised ?Concept.}{?AyatNo qreg:hasPart ?Concept.} OPTIONAL {?AyatNo rdfs:comment ?Ayat.}}
**Answer:** The answer of the above query is shown in Figure 4.

Fig. 4. Result of SPARQL query 2.

The verse is retrieved along with verse number, whole verse is not shown due to space constraints in the page.

**Query 3**: Which verses contain concepts Allah and Earth?
**SPARQL Query 3.** SELECT * WHERE {{?AyatNo qreg:hasPart qreg:Allah.} {?Ayat qreg:hasPart qreg:Earth.} *OPTIONAL { ?AyatNo rdfs:comment ?Ayat.}}*

**Answer:** The answer of the above query is shown in Figure 5.

| AyatNo | |
|---|---|
| 2:107 | "Do you not know that to Allah bel |
| 229 | "It is He who created for you all of |
| 2:116 | "They say, " Allah has taken a son |
| 2:164 | "Indeed, in the creation of the hea |
| 230 | "And [mention, O Muhammad], wh |
| 222 | "He who made for you the earth a |
| 2:61 | "And [recall] when you said, "O Mo |
| 2:117 | "Originator of the heavens and the |

Fig. 5. Result of SPARQL query 3.

**Query 4:** Name the animal which swallowed Prophet Yunus (Jonah)?
**SPARQL Query 4.** SELECT ?Answer WHERE {?Answer qreg:swallowed qreg:Yunus.}
**Answer:** Fish.

## V. CONCLUSION AND FUTURE WORK

The search for particular concepts of the holy Quran and the retrieval of the verses containing those concepts is an extremely significant problem for people who are curious in gaining Quranic knowledge. The nature of the problem is not too different for the Bible, the Torah and other religious texts. People of all faiths often search for concepts of interest, verse number and text of the verse containing those concepts on popular web search engines. With keyword based search engines, users need to remember many or most keywords of a particular verse in order to get a desired search result. Ontological modeling of the holy Quran (or any other religious text) offers an alternative notion for search, "search based on the concepts of the verse and use of the relation between concepts in the search procedure". The inspiration for this kind of search and modeling has been drawn from the success of semantic web technologies. In fact, the very essence of semantic web is semantics/concept based search. The preliminary stage for such a search mechanism requires an ontological modeling of the data. In this research, we have done just that. We have made an ontological model of the nature related concepts described in the Quranic verses which are spread out over different chapters. Later, SPARQL queries have been used to retrieve concepts of interest, the full text of the verse and the verse number. Our modeling ensures that similar/same concepts of the holy Quran which are spread out over different chapters and verses can be retrieved using single queries. This indeed paves the way for an improvement over existing search mechanisms because the verse retrieval is possible if the user remembers only the main concepts of a verse rather than many keywords. In addition, concept retrieval is possible through use of relations.

While some other research works described in this paper apply ontological modeling and semantic retrieval of Quranic concepts, our paper is different because it exclusively focuses on the "nature" domain of the holy Quran. In addition, storage and retrieval mechanism for the full text and the verse number of the Quranic verse is different in our research and effective as well. We strongly believe that, it is a major step towards semantic search of nature related concepts of the holy Quran. We are working to build a public SPARQL endpoint so that inquisitive web users can use the power of semantic search to query about nature related concepts of the holy Quran.

REFERENCES

[1] I. Horrocks, "Ontologies and the Semantic Web," *Communications of the ACM*, Volume 51 Issue 12, December 2008.
[2] Noorhan Hassan Abbas, "Quran 'Search for a Concept' Tool and Website (Master's thesis)," School of Computing, University of Leeds, England, 2009.
[3] MA Sherif, and Axel-Cyrille Ngonga Ngomo, "Semantic Quran," Semantic Web, vol. 6, no. 4, 7 August 2015.
[4] Tanzil Project. (n.d.). Retrieved December 5, 2014, available at http://tanzil.net/wiki/Tanzil_Project
[5] Ontology of Quran Concept: Available at http://corpus.quran.com/ontology.jsp. Leeds University, UK.
[6] Aya M. Al-Zoghby, Ahmed SharafEldin Ahmed and Taher T. Hamza, "Arabic Semantic Web Applications – A Survey," *Journal of Emerging Technologies in Web Intelligence, Vol. 5, No. 1 (2013), 52-69, Feb 2013*.
[7] H.U. Khan, S. M. Saqlain, M. Shoaib, and M. Sher, "Ontology Based Semantic Search in Holy Quran," *International Journal of Future Computer and Communication*, Vol. 2, No. 6, December 2013.
[8] S. Saad, and N. Salim, "Build Islamic Ontology based on Ontology Learning," *Postgraduate Annual Research Seminar, UTM, Johor Bahru, Malaysia ,Vol. 59, 2007*.
[9] S. Saad, N. Salim, and N. Omar, "Keyphrase Extraction for Islamic Knowledge Ontology," Information Technology, 2008, ITSim 2008, International Symposium on. Vol. 2, IEEE, 2008.
[10] S. Saad, N. Salim, and H. Zainal, "Towards Context-sensitive Domain of Islamic Knowledge Ontology Extraction," *International Journal for Infonomics (IJI) Volume 3, Issue 1, March 2010*.
[11] A. B. M. Sharaf, and E. Atwell, "QurSim: A Corpus for Evaluation of Relatedness in Short Texts," *In LREC* (pp. 2295-2302), Istanbul, Turkey.
[12] Z. A. Adhoni, and A. A. Siddiqi, "A Programming Approach for the Digital Quran Applications," *International Journal of Engineering & Computer Science IJECS-IJENS, Vol. 13 No. 05, October 2013*.
[13] J. Dror, D. Shaharabani, R. Talmon, and S. Wintner, "Morphological Analysis of the Qur'an," *Literary and Linguistic Computing*, 19(4), 431-452.
[14] P. Saeedi, S. Heidari, and M. Farhoodi, "Creating Quranic Question Taxonomy," *Electrical Engineering (ICEE), 2014 22nd Iranian Conference on*. IEEE, 2014.

[15] Yauri, Aliyu Rufai, "Quranic Verse Extraction base on Concepts Using OWL-DL Ontology," *Research Journal of Applied Sciences, Engineering and Technology* 6.23 (2013): 4492-4498.

[16] Q. U. Ain, and A. Basharat, "Ontology Driven Information Extraction from the Holy Qur'an Related Documents,"*26th IEEEP Students' Seminar*, 2011.

[17] Azman Ta'a ,Syuhada Zainal Abidin, Mohd Syazwan Abdullah, Abdul Bashah B Mat Ali, and Muhammad Ahmad, "AL-QURAN Themes Classification using Ontology," Proceedings of the 4[th] International Conference on Computing and Informatics (ICOCI 2013), 28-30 August, 2013 Sarawak, Malaysia.

[18] Saheeh International: The Saheeh International™ Team & Dar Abul-Qasim. (n.d.), available at http://www.saheehinternational.com/

[19] M. Horridge, S. Jupp, G. Moulton, A. Rector, R. Stevens, C. Wroe, "A Practical Guide to Building OWL Ontologies using Protégé 4 and CO-ODE Tools, Edition 1.2," The University of Manchester, 2009.

[20] N. Drummond, M. Horridge, H. Knublauch, Protégé-OWL Tutorial. In 8th International Protégé Conference, Madrid, Spain.

[21] G. Antoniou, and F. V. Harmelen, "A Semantic Web Primer," MIT press, 2004.

[22] I. Kollia, B. Glimm, I. Horrocks, "SPARQL Query Answering over OWL Ontologies," Proceedings of the 8th Extended Semantic Web Conference on the Semantic Web: Research and Applications, May 29-June 02, 2011, Heraklion, Crete, Greece